\documentclass[letterpaper, 10 pt, conference]{ieeeconf}  

\IEEEoverridecommandlockouts                              

\usepackage{graphics} 
\usepackage{epsfig} 
\usepackage{mathptmx} 
\usepackage{times} 
\usepackage{amsmath} 
\usepackage{amssymb}  
\usepackage{algorithm}
\usepackage{algorithmic}
\usepackage{newfloat}
\usepackage{listings}
\usepackage{booktabs}

\usepackage{times}  
\usepackage{helvet}  
\usepackage{courier}  
\usepackage[hyphens]{url}  
\usepackage{graphicx} 
\usepackage{caption}
\usepackage{bibentry}
\usepackage{todonotes}
\usepackage{hyperref}
\usepackage{makecell}
\usepackage{float}

\makeatletter
\newcommand\fs@ruledalgdown{%
  \def\@fs@cfont{\bfseries}%
  \let\@fs@capt\floatc@ruled
  \def\@fs@pre{\vspace*{1.46mm}\hrule height.8pt depth0pt \kern2pt}%
  \def\@fs@mid{\kern2pt\hrule\kern2pt}%
  \def\@fs@post{\kern2pt\hrule\relax}%
  \let\@fs@iftopcapt\iftrue
}
\floatstyle{ruledalgdown}
\restylefloat{algorithm}
\makeatother

\title{\LARGE \bf Novelty Adaptation Through Hybrid Large Language Model (LLM)-Symbolic Planning and LLM-guided Reinforcement Learning}

\author{Hong Lu$^{1}$, Pierrick Lorang$^{1,2}$, Timothy R. Duggan$^{1}$, Jivko Sinapov$^{1}$, Matthias Scheutz$^{1}$%
\thanks{*This work has been in part funded by ONR grant N00014-25-1-2479.}%
\\ [1ex]
$^{1}$ Department of Computer Science, Tufts University, Medford, MA, USA \\
$^{2}$ Austrian Institute of Technology GmbH, Vienna, Austria
}

\begin{document}

\maketitle
\thispagestyle{empty}
\pagestyle{empty}

\begin{abstract}

In dynamic open-world environments, autonomous agents often encounter novelties that hinder their ability to find plans to achieve their goals. Specifically, traditional symbolic planners fail when the robot's planning domain lacks the operators to interact appropriately with novel objects. We propose a neuro-symbolic architecture that integrates symbolic planning, reinforcement learning, and a large language model (LLM) to learn how to handle novel objects. In particular, we leverage the common sense reasoning capability of the LLM to identify missing operators, generate plans with the symbolic AI planner, and write reward functions to guide the reinforcement learning agent in learning control policies for newly identified operators. Our method outperforms the state-of-the-art methods in operator discovery as well as operator learning in continuous robotic domains. Our webpage and code can be access here: \href{https://helenlu66.github.io/hybridLLMguided/}{helenlu66.github.io/hybridLLMguided/}

\end{abstract}

\section{Introduction}
Integration of high-level task reasoning with low-level motion
planning within a Task and Motion Planning (TAMP) framework improves
the ability of a robotic system to solve complex problems. Using its
knowledge of the environment and task structure encoded in the Planning Domain Definition Language (PDDL)
\cite{aeronautiques1998pddl}, a high-level symbolic task planner can
generate a sequence of operator executions using the
operator's pre- and post-conditions to achieve the desired goal
\cite{ghallab_automated_2016} while a low-level motion planner can
generate the robot motion for each operator execution
(e.g., \cite{disprod23}).
Alternatively, each operator could have a policy that was
learned through Reinforcement Learning (RL) (e.g.,
\cite{goel_rapid-learn_2022}).

\begin{figure*}[th]
    \centering 
    \includegraphics[width=1.0\linewidth]{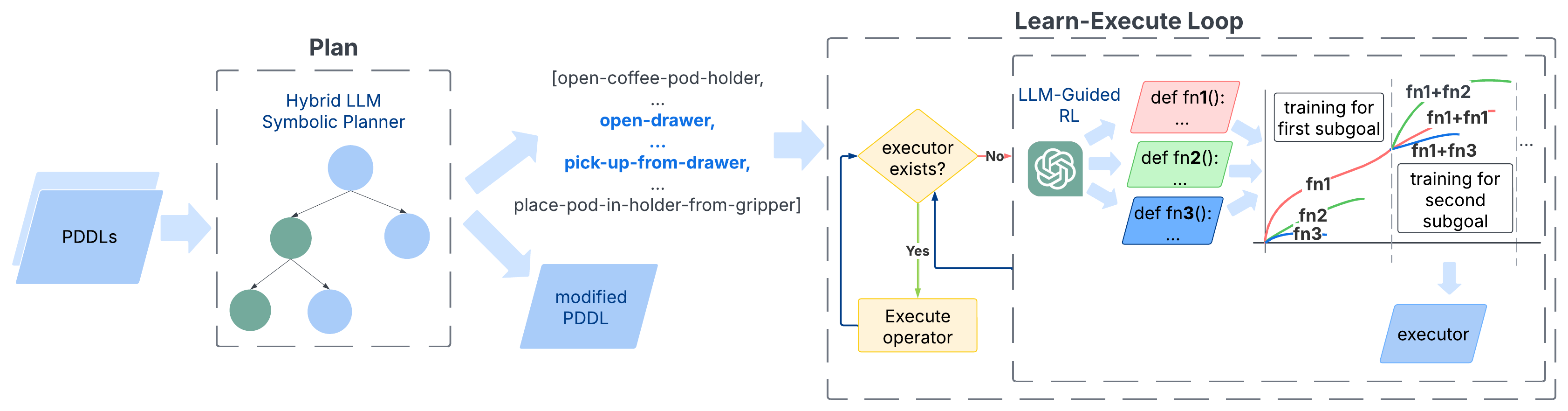}
    \caption{The plan-learn-execute loop. The Hybrid LLM Symbolic planner parses the domain and problem PDDL files, prompts the LLM to define the missing operator(s) in PDDL, and finds a plan with grounded operators. It lifts the grounded operators, and outputs the modified domain PDDL file with the added lifted operator definitions and the plan. The LLM's newly defined operators are in blue while the existing ones are in black. The learn-execute loop starts by executing the operators in the plan. When it encounters a new operator for which an executor policy does not yet exist, it prompts the LLM to generate dense reward shaping function candidates and launches RL agents to learn the policy. The effects of the operator are treated as sub-goals for the RL agents. One agent is launched per dense reward function candidate and the worst performing agent is eliminated periodically based on sub-goal success rates. The sub-goals are trained in phases. Once training is complete, the best performing policy is saved as the operator's executor. The pseudocode for the algorithm is shown in Algorithm \ref{alg:plan_learn_execute}.}
    \label{fig:plan-learn-execute}
\end{figure*}

However, planning failures can occur due to novelties when the robot
cannot find an appropriate sequence of operators to achieve its goal
because the solution is outside the robot's planning domain
\cite{sarathy_spotter_2020, Gizzi_2022}.  In fact, common everyday objects,
when introduced into a robot's controlled environment, often disrupt
the robot's ability to achieve its usual goals because the robot's
predefined planning domain lacks the operators necessary for
interacting with these objects. In this work, we focus on the introduction of common objects and the challenge of adapting to them, i.e., for the robot to learn how to manipulate them.  As such objects
might appear within the training data of LLMs, an LLM could
effectively reason about interactions involving them
\cite{zhao2023largelanguagemodelscommonsense}. We thus utilize an LLM to perform
common sense reasoning to structurally define missing plan operators in the PDDL format. In addition, to
efficiently guide the exploration of the RL algorithm for learning the
associated operator policies in a continuous space, we prompt
the LLM to write dense reward function candidates that are used in
a sub-goal curriculum. We launch multiple RL agents---one per
dense reward function candidate for each sub-goal---and periodically
prune the worst performing candidate until a policy is found that can
generate robot motion from the operator's pre-conditions to its
post-conditions. Figure~\ref{fig:plan-learn-execute} illustrates our
approach. Our contributions are as follows:

\begin{enumerate}
    \item We propose a method that leverages the common-sense reasoning of LLMs to solve the missing operator identification problem—a challenge that is intractable for traditional operator discovery methods in the hybrid planning and learning literature.
    \item We augment a hybrid planning and learning architecture with LLM generated dense reward functions, thereby significantly speeding up the learning process. 
    \item We perform evaluations in continuous robotic manipulation domains in the mimicgen simulation environment \cite{mandlekar_mimicgen_2023}.
\end{enumerate} 

\section{Related Works}
\subsection{Hybrid Planning and Reinforcement Learning}
To handle the novelties that might arise in an agent's environment, Goel et al. and Sarathy et al. propose hybrid planning and learning frameworks that extend the definition of MacGyver problems \cite {sarathy_spotter_2020, goel_rapid-learn_2022, goel_neurosymbolic_2024}. In these frameworks, when the agent fails to plan due to novelties, RL agents are spawned to bridge the gap by discovering plannable states from which the goal is reachable. In addition, the structured knowledge in the symbolic reasoner provides guidance to the RL agent during its exploration \cite{yang_peorl_2018,kokel_reprel_2021,acharya_neurosymbolic_2024,balloch_neuro-symbolic_2023, illanes_symbolic_2020, guan_leveraging_2022}. However, these works often evaluate their approaches in discrete action domains. Recent works in robotics have applied hybrid planning and learning methods to continuous domains \cite{silver_learning_2022,lorang_framework_2024,cheng_league_2023,lorang_adapting_2024}. Notably, LEAGUE~\cite{cheng_league_2023} integrates task planning with skill learning for long-horizon manipulation. 
These works typically rely on RL agents to discover the missing operator or recover from operator 
failure, spending considerable time during exploration. To structure this exploration, reward 
machine-based approaches have proposed decomposing the reward function according to task 
structure \cite{icarte2022reward, lorang_curiosity-driven_2025}. Nevertheless, even when guided with symbols, these works never explicitly reason about missing operators, instead relying on RL agents to discover them incidentally through exploration, a strategy that scales poorly in continuous domains where plannable states are difficult to reach by chance. In this work, we explore whether the common sense encoded in LLMs can be utilized to structurally define the missing operators and guide the RL agents via reward shaping, thereby saving exploration time. 
\begin{figure*}[th]
    \centering 
    \includegraphics[width=1\linewidth]{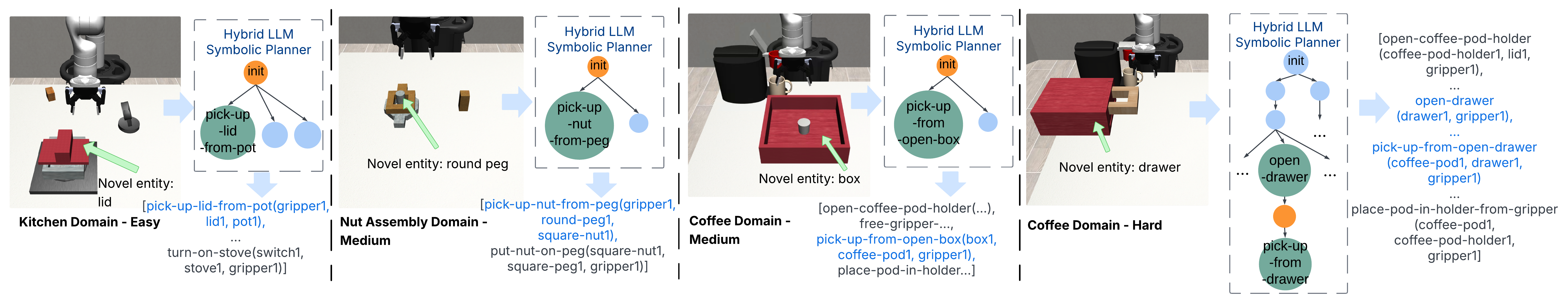}
    \caption{Problem domains and hybrid LLM-symbolic planner outputs. Domains are ordered by the difficulty of discovering a plannable state via random exploration. Green arrows indicate injected novelties: a lid (Kitchen), a round peg (Nut Assembly), and a drawer or box (Coffee). In the planning graph, green nodes represent states reached via LLM-suggested operators, blue nodes denote existing operators, and orange nodes indicate where the search-ahead algorithm finds a valid plan. Identified missing operators (shown in blue text) include: \textit{pick-up-lid-from-pot} (Kitchen), \textit{pick-up-nut-from-peg} (Nut Assembly), \textit{pick-up-from-open-box} (Coffee-Box), and \textit{open-drawer} and \textit{pick-up-from-open-drawer} (Coffee-Drawer).}
    \label{fig:hybrid planner}
\end{figure*}
\subsection{LLM for Planning}
Recent works in robotics planning seek to leverage the semantic understanding, common sense knowledge and the code generation capability of LLMs. For example, Silver et al. propose a prompting pipeline and evaluate LLMs' ability to write code that solves general problems in a given PDDL domain \cite{silver_generalized_2024}, Singh et al. prompt LLMs to write functions that call available robot actions to achieve a goal \cite{singh_progprompt_2022}. Some works aim to increase the executability of the LLM's proposed plan by grounding the plan in the robot's executable actions \cite{huang_language_2022,ahn_as_2022,huang_grounded_2023,singh_progprompt_2022}. To plan in dynamic and partially observable environments, some works introduce interactive methods of prompting the LLMs to incorporate environment feedback and self-correct \cite{yao_react_2023, huang_inner_2022, dagan_dynamic_2023, wang_describe_2024}. Hybrid approaches combine the capabilities of the LLMs with the reliability of symbolic planners to overcome the lack of flexibility of symbolic planners and the tendency to hallucinate of the LLMs. For example, Liu et al. leverage the LLMs' natural language understanding to translate natural language problems into a PDDL problem based on the given PDDL domain and use a symbolic planner to generate the plan \cite{liu_llmp_2023}, and Dagan et al. leverage the common sense encoded in LLMs to make assumptions about the truth value of predicates in partially observable environments \cite{dagan_dynamic_2023}. A common assumption made in these works is that the existing operators are enough to reach the goal state. Unlike these prior works that focus on plan synthesis, our approach focuses on domain expansion and leverages the LLM to infer the missing operator(s) when a novel object is injected. 
\vspace{-0.1cm}
\subsection{LLM for Reinforcement Learning}
LLMs can be used to either suggest goals for exploration or provide reward signals. For example, LLMs can suggest the next goal to explore when the RL agent is faced with a new task \cite{du_guiding_2023}, suggest sequences of sub-goals as potential paths to the goal \cite{shukla_lgts_2024}, or suggest auxiliary tasks for learning to maximize the use of previous experiences collected during RL training \cite{quartey_exploiting_2024}. In addition, LLMs can provide reward signals either by writing reward function codes or by directly evaluating an episode's outcome \cite{kwon_reward_2023, song_self-refined_2023, xie_text2reward_2024, ma_eureka_2024}. Closest to our work is LEAGUE++~\cite{li2024league++}, which shares our motivation 
of leveraging LLM-generated reward functions to guide the acquisition of new robot 
skills. Despite this conceptual proximity, two important limitations set it apart 
from our approach. First, LEAGUE++ assumes a fully specified planning domain, meaning 
it has no mechanism to reason about or recover from the absence of operators needed 
to handle novel objects --- a critical shortcoming for open-world deployment. Second, 
it commits to a single LLM-generated reward function per skill, leaving the agent exposed to the risk of investing substantial 
training time following a fundamentally misguided signal. Our work addresses both 
gaps: we introduce an LLM-guided symbolic search to explicitly identify missing 
operators, decompose their effects into an ordered sub-goal curriculum, and generate 
multiple reward function candidates in parallel, applying a genetic algorithm-inspired 
elimination strategy to progressively retain the most effective reward signal 
throughout training.
\subsection{Prompting Techniques for Reasoning}
Wei et al. demonstrate that reasoning can be elicited from LLMs when they emulate the chain-of-thought process shown in their prompts \cite{wei_chain--thought_2023}. Kojima et al. point out that similar thought process can be elicited by simply adding ``let's think step by step'' to the prompt \cite{kojima_large_2023}. The self-consistency approach demonstrates that the accuracy of the output can be further improved if multiple chains-of-thoughts are sampled and the most frequent output is chosen \cite{wang_self-consistency_2023}. Tree-of-thoughts (ToT) takes this one step further by prompting the LLM to generate multiple next states in a given problem state and evaluate the progress each thought makes. Inspired by the exploratory nature of ToT, we employ self-consistency prompting to sample the most frequently proposed missing operator in each planning state, and we use the symbolic planner as a binary state evaluator that searches ahead and reports whether a plan can be found with the added operator.

\section{Preliminaries}

\subsection{Symbolic Planning} Given a formal definition of a problem domain $\sigma = \langle \mathcal{E}, \mathcal{F}, \mathcal{S}, \mathcal{O}\rangle$, where $\mathcal{E}$ is a set of typed entities in the environment, $\mathcal{F}$ a set of detectable Boolean predicates applicable to entities, $\mathcal{S}$ a set of states represented as a set of grounded predicates, and $\mathcal{O}$ a set of operators, we seek to find a solution $\mathcal{P}=[o_1,\ldots,o_{|\mathcal{P}|}]$ for the planning task $\mathcal{T}=(\mathcal{E}, \mathcal{F}, \mathcal{O}, s_0, s_g)$.  $\mathcal{P}$  is a sequence of grounded operators that when applied transforms the initial state $s_0$ into the goal state $s_g$. Each operator $o \in \mathcal{O}$ is defined as the entities it operates on, the preconditions $\psi$ that must be satisfied before it can be applied, and the effects $\omega$ that become true once it is applied~\cite{aeronautiques1998pddl}.

\subsection{Neuro-symbolic Planning and Control} Neuro-symbolic planning and control retain the symbolic planning problem formulation above. For each operator $o_i \in \mathcal{P}=[o_1,\ldots,o_{|\mathcal{P}|}]$, a neural policy $\pi_i \in \Pi$ is learned. Each skill $\pi_i$ is executable when the preconditions $\psi_i$ of the operator are met, and interacts with the environment to realize the operator's effects $\omega_i$. This formulation captures the learning of low-level controls and the abstraction of tasks in high-level planning.

\subsection{Problem Formulation} Given a planning task $\mathcal{T'}=(\mathcal{E'}, \mathcal{F}, \mathcal{O}, s_0, s_g)$ after injection of a novel entity $e$ where $\mathcal{E'} = \mathcal{E} \cup e$, obtain an augmented set of operators $\mathcal{O'}$ by adding the missing operators $\mathcal{O}_{missing}$ that can be applied to $e$ so that a plan $\mathcal{P'}$ can be found from the initial state $s_0$ to the goal state $s_g$. Furthermore, learn the neural skill $\pi_i$ associated with each new grounded operator $o_i \in \mathcal{O}_{missing}$. We assume that the available predicates are enough to define the missing operators and the robot is able to detect and categorize the novel object. For example, since the robot can detect the truth value of the predicate \textit{open(container)}, we assume that it can recognize that a drawer is a container type and detect whether the drawer is open after it is injected as a novelty.
\section{The Plan-Learn-Execute Loop}
\begin{figure*}[th]
    \centering 
    \includegraphics[width=1.0\linewidth]{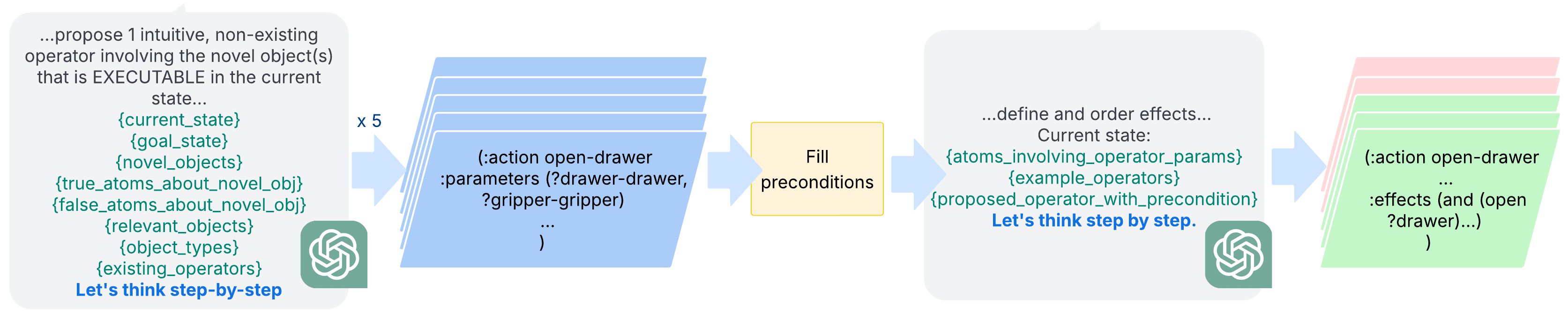}
    \caption{The Prompt-LLM-for-New-Operator Pipeline. We sample five operator candidates and select the majority \cite{wang_self-consistency_2023}. The prompt is dynamically filled in with the current state, goal, and existing operators. The LLM suggests names and parameters for missing operators. Preconditions are automatically filled with grounded predicates involving these parameters, while the LLM defines and orders the effects. For example, if preconditions for \textit{open-drawer} include \textit{not open drawer1} and \textit{not grasped drawer1}, the LLM generates \textit{grasped drawer1} and \textit{open drawer1} as ordered effects, which subsequently serve as sub-goals for the guided learning stage. We observe that errors in effects ordering and operator generation are greatly reduced through self-consistency \cite{wang_self-consistency_2023}. Finally, the grounded operator is lifted into a general PDDL definition by mapping specific entities to their variable types (e.g., drawer1 to ?d - drawer), enabling domain-wide generalization.}
    \label{fig:operator prompting}
\end{figure*}
The loop of plan-learn-execute shown in Figure \ref{fig:plan-learn-execute} and Algorithm \ref{alg:plan_learn_execute} begins by calling the search-and-prompt algorithm of the hybrid planner to solve the missing operator identification problem. The search and prompt algorithm augments the problem domain PDDL with definitions of $O_{missing}$ and returns a plan that contains grounded operators of $O_{missing}$ for which the robot does not know how to execute. The plan-learn-execute algorithm learns the neural control policy for each operator in $O_{missing}$ through the LLM guided sub-goal learning process.

\begin{algorithm}[t]
\footnotesize
\caption{\textit{\textbf{Plan-Learn-Execute}}}
\label{alg:plan_learn_execute}
\begin{algorithmic}[1]
\STATE \textbf{Input:} $\mathcal{T'}$, $max\_iter$
\STATE \textbf{Output:} $goal\_achieved$
\STATE $iteration \leftarrow 0$
\STATE $goal\_achieved \leftarrow \FALSE$
\WHILE{$iteration < max\_iter$ \textbf{and} $\neg goal\_achieved$}
    \STATE \textbf{ResetEnvironment}()
    \STATE $\mathcal{P'} \leftarrow$ \textbf{Search-And-Prompt}($\mathcal{T'}$)
    \WHILE{$|\mathcal{P'}| > 0$}
        \STATE $o \leftarrow$ \textbf{PopFirst}($\mathcal{P'}$)
        \STATE $success, executor \leftarrow$ \textbf{ExecuteOperator}($o$)
         \IF{$\neg executor$}
            \STATE $executor \leftarrow$ \textbf{LLMGuidedSubgoalLearning}($o$)
            \STATE $success, executor  \leftarrow$ \textbf{ExecuteOperator}($o$)
        \ENDIF
        \IF{$\neg success$}
            \STATE \textbf{break} \COMMENT{restart the main loop}
        \ENDIF
    \ENDWHILE
    \STATE $iteration \leftarrow iteration + 1$
    \STATE $goal\_achieved \leftarrow |\mathcal{P}| = 0 \textbf{ and } success$
\ENDWHILE
\RETURN $goal\_achieved$
\end{algorithmic}
\end{algorithm}

\subsection{Hybrid LLM and Symbolic Planning}
\begin{algorithm}[th]
\footnotesize
\caption{\textit{\textbf{Search-and-Prompt}}}
\label{alg:search-and-prompt}
\begin{algorithmic}[1]
\STATE \textbf{Input: }$\mathcal{T'}$, $max\_depth$
\STATE \textbf{Output:} $\mathcal{O'}=\mathcal{O}\cup \mathcal{O}_{missing}, \mathcal{P'}$
\STATE $open\_queue \leftarrow [s_0, \mathcal{O})]$
\STATE $closed \leftarrow [(s_0, \mathcal{O})]$
\STATE $depth \leftarrow 0$
\WHILE{$|open\_queue|>0$}
    \STATE $s, \mathcal{O} \leftarrow$ \textbf{PopFirst}($open\_queue$)
    \STATE $o_{missing} \leftarrow$ \textbf{Prompt-LLM-For-New-Operator}($s, \mathcal{O}$)
    \STATE $\mathcal{O'} \leftarrow \mathcal{O} \cup o_{missing}$
    \STATE $\mathcal{P'} \leftarrow$ \textbf{Search-Ahead}($s$, $\mathcal{O'}$, $max\_depth$) \COMMENT{search symbolically to reduce LLM calls}
    \IF{$\mathcal{P'} \neq $ None}
            \RETURN $\mathcal{O', P'}$
    \ENDIF
    \IF{$depth \geq max\_depth$}
            \STATE break
    \ENDIF
    \STATE $\mathcal{S'} \leftarrow $ \textbf{Applicable}($s, \mathcal{O'}$)
    \FOR {$s' \in \mathcal{S'}$}
            \IF{$s' \notin closed$}
                    \STATE $open\_queue \leftarrow open\_queue \cup [(s', \mathcal{O'})]$
                    \STATE $closed \leftarrow closed \cup [(s', \mathcal{O'})]$
            \ENDIF
    \ENDFOR
    \STATE $depth \leftarrow depth + 1$
\ENDWHILE
\RETURN None, None
\end{algorithmic}
\end{algorithm}
\begin{algorithm}[th]
\footnotesize
\caption{\textit{\textbf{LLM-sub-goal-Reward}}}
\label{alg:llm_subgoal_reward}
\begin{algorithmic}[1]
\STATE \textbf{Input: }o, prev\_num\_obs, curr\_num\_obs, curr\_binary\_obs
\STATE \textbf{Output:} reward
\STATE reward $\leftarrow$ 0
\FOR{$i = 1 \to$ $|$o.effects$|$}
    \STATE sub-goal\_bonus $\leftarrow 10^i$
    \STATE max\_reward\_shaping $\leftarrow 10^{i-1}$
    \STATE effect $\leftarrow$ o.effects[i]
    \IF{$\neg$\textbf{CurrTrainingPhase}(effect)}
        \STATE \textbf{break} \COMMENT{avoid giving rewards for locked sub-goals}
    \ENDIF
    \IF{\textbf{CheckEffectSatisfied}(effect, curr\_binary\_obs)}
        \STATE reward$\leftarrow$ reward + \textit{ \textit{sub-goal\_bonus}}
    \ELSE 
        \STATE reward $\leftarrow$ reward + max\_reward\_shaping $\times$ \textbf{LLM-Reward-Shaping}(prev\_num\_obs, curr\_num\_obs,effect)
        \STATE break \COMMENT{stop as soon as a sub-goal is not satisfied}
    \ENDIF
\ENDFOR
\RETURN reward
\end{algorithmic}
\end{algorithm}
The Hybrid LLM Symbolic Planner takes a PDDL domain specifying the available operators $\mathcal{O}$, predicates $\mathcal{F}$, object types and a PDDL problem file specifying the set of entities $\mathcal{E'}$ including the novel entity $e$, the initial state $s_0$ and the goal state $s_g$ as grounded predicates over entities. The hybrid planner goes through a breadth-first search-and-prompt process to find a plan to the goal as illustrated by Figure \ref{fig:hybrid planner} and Algorithm \ref{alg:search-and-prompt}. 
\begin{figure*}[th]
    \centering 
    \includegraphics[width=1.0\linewidth]{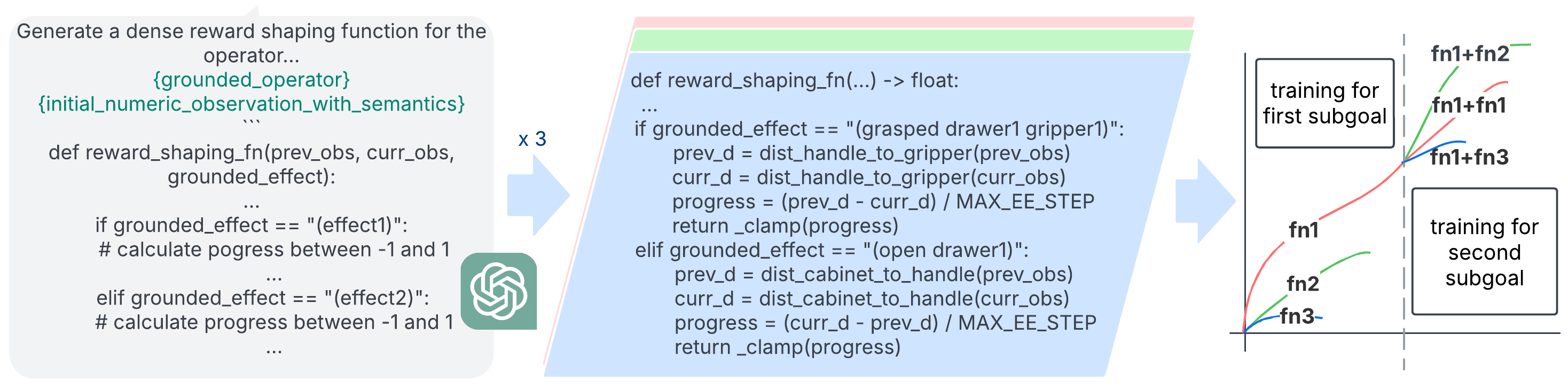}
    \caption{The LLM Guided Sub-goal Learning Pipeline. Three reward shaping function candidates are sampled from the LLM. We use a temperature of 1.0 to encourage diversity in the samples. We filter out semantically identical candidates by passing in test values to the candidates and eliminating duplicates that output the same reward. The prompt contains the template for a reward shaping function. The definition of the grounded operator and the observation space of the robot are dynamically filled into the prompt. The LLM writes a function to compute a velocity based progress using the relevant observations in the robot's observation space. For example, the function computes the progress for \textit{open drawer1} using the distance between the drawer handle and the cabinet. During each sub-goal's training phase, the reward shaping for the sub-goal is unlocked. The worst performing candidate is periodically eliminated.}
    \label{fig:reward shaping prompting}
\end{figure*}
In the search-and-prompt algorithm, the LLM is prompted in each search state to suggest up to one missing executable operator shown in Figure \ref{fig:operator prompting}. We employ the self-consistency prompting technique to increase the accuracy of the output \cite{wang_self-consistency_2023}. To minimize LLM calls, a purely symbolic search-ahead algorithm evaluates whether a plan can already be found from the current state forward with the addition of the suggested operator. The search-ahead algorithm is a breadth-first search that searches for the goal state given the available operators. If no plan can be found, the operators whose preconditions are satisfied in the current state are applied to reach next states, and the reachable next states are added to the search-and-prompt tree. The search-ahead algorithm terminates the search-and-prompt process early if a plan with the currently available operators is found. The search-and-prompt tree grows until a plan is found or the max search depth is reached. We choose GPT-o3 for the planning stage for its advanced reasoning capabilities.
\section{LLM Guided Sub-goal Learning}

We launch Proximal Policy Optimization (PPO) agents to learn the continuous control for each grounded operator in the plan \cite{schulman2017proximalpolicyoptimizationalgorithms}. Our RL pipeline is implemented using the \textit{stable\_baselines3}~\cite{stable-baselines3} library. We guide the PPO agents toward each sub-goal via reward shaping functions written by GPT-o4-mini with a temperature of 0.3. We choose GPT-o4-mini instead of GPT-o3 for the function writing due to its fast and cost-efficient code generation. We employ a genetic algorithm inspired approach to filter the candidates by eliminating the non-runnable ones and iteratively eliminating the least successful ones.

Figure \ref{fig:reward shaping prompting} illustrates the generation and filtering of reward shaping functions. The robot receives numeric observations (e.g., relative distances between the gripper and task-relevant entities) and binary observations (grounded predicate truth values). These, along with the operator definition and maximum end-effector displacement, are used to prompt the LLM to generate three reward function candidates. Each candidate outputs a linear, velocity-based reward in the range $[-1, 1]$ based on the change in numeric observations between timesteps. For example, in the \textit{open-drawer} task (Fig. \ref{fig:reward shaping prompting}), the LLM-generated code calculates and normalizes the distance change between the cabinet and handle, returning a clamped progress reward towards the \textit{open-drawer} sub-goal.
\begin{figure*}[th]
    \centering 
    \includegraphics[width=0.92\linewidth]{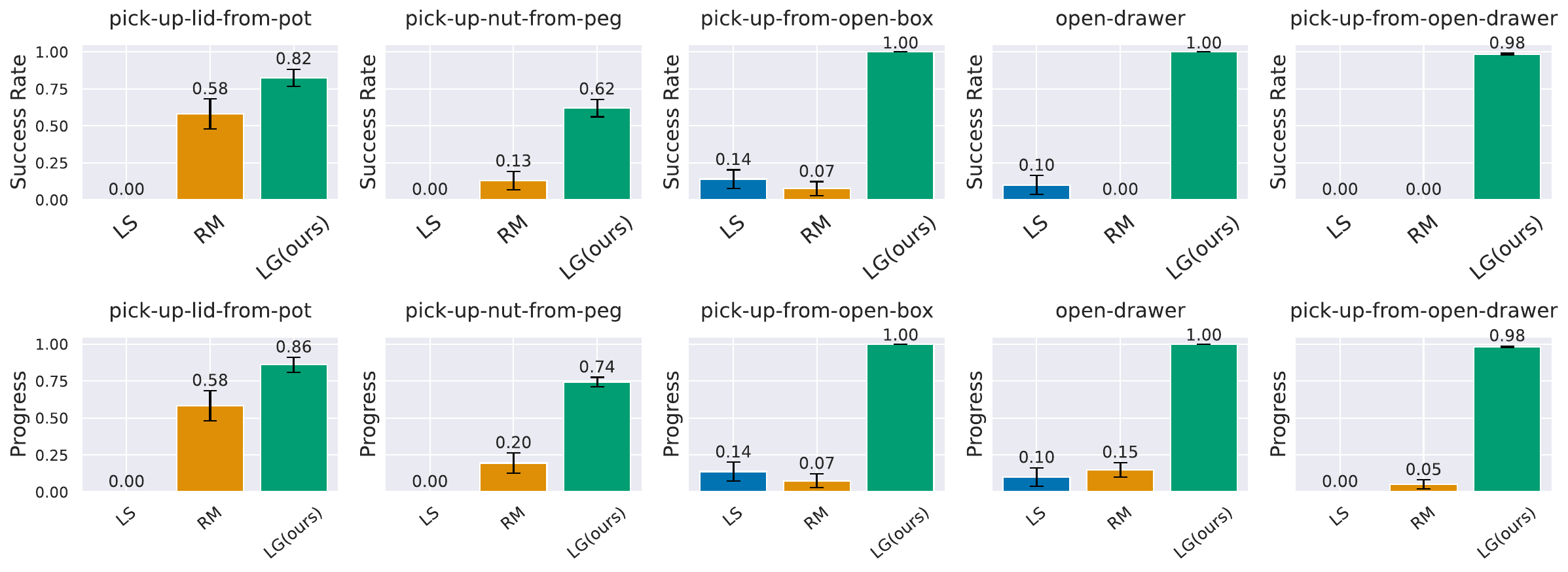}
    \caption{Comparison of the LLM Guided (LG) Sub-goal Learning with the LEAGUE-Sparse (LS) \cite{cheng_league_2023} and Reward Machine (RM)  \cite{icarte2022reward} Baselines. We record two metrics: success rate of the operator and progress towards the completion. Progress is the percent of sub-goals achieved. Metrics are averaged across ten seeds with standard error of the mean (SEM).}
    \label{fig:sr pr barplots}
\end{figure*}
\begin{figure*}
    \centering
    \includegraphics[width=1.0\linewidth]{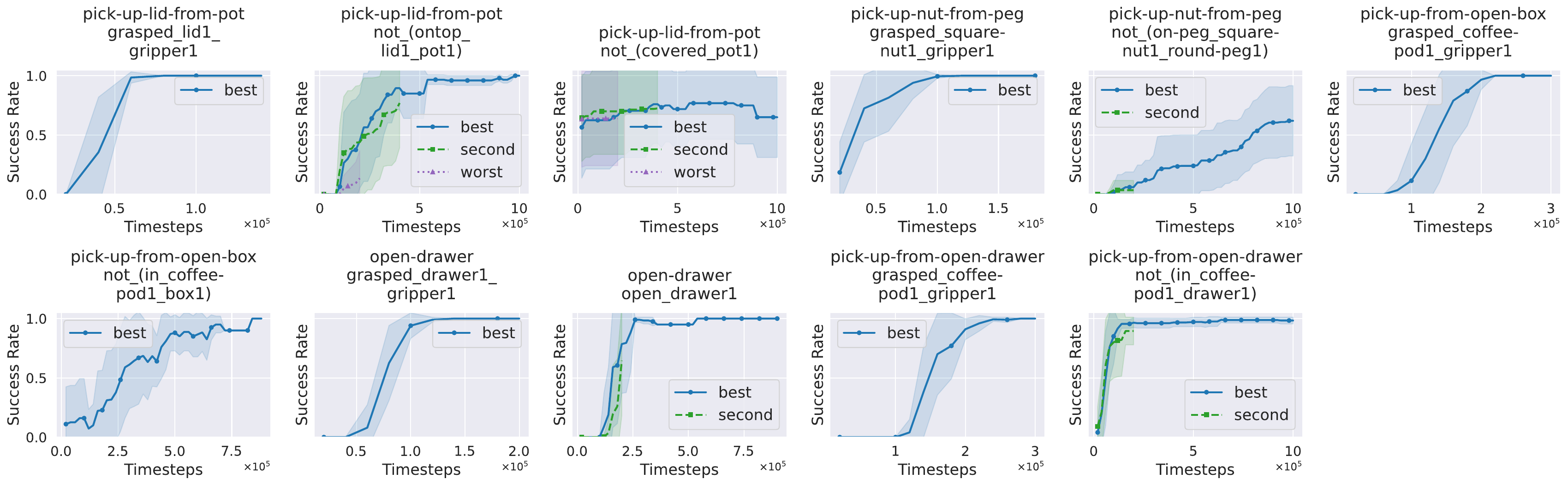}
    \caption{Reward Function Candidates Evaluation Curves For Each Sub-goal During Training. The worst performing candidate was eliminated at $2 \times 10^5$ timestep intervals. Note that some sub-goals did not have three reward function candidates due to them being semantically identical. Refer to Table~\ref{tab:reward_diversity} for the number of unique reward function candidates per sub-goal.}
    \label{fig:eval_curves}
\end{figure*}

\begin{table}[th]
\centering
\caption{Hybrid LLM Symbolic Planner vs Operator Discovery (OD)}
\label{tab:planner vs rl}
\begin{tabular}{@{}lrrrr@{}}
    \toprule
    &Hybrid & OD& Hybrid & OD\\
    Domain &  Successes & Successes &Time & Time\\
    \cmidrule[0.4pt]{1-5}
    Kitchen &  $10/10$ & $10/10$ & 75.19 sec & 85.51 sec  \\
    Nut Assembly &  $10/10$ & $0/10$ & 120.76 sec & $>7$h \\
    Coffee (box) & $10/10$ & $0/10$ & 45.26 sec &  $>7$h\\
    Coffee (drawer)&  $7/10$ & $0/10$ & 681.83 sec &  $>7$h\\
    \bottomrule
\end{tabular}
\end{table}

As effects i.e. sub-goals of an LLM defined operator in Figure~\ref{fig:operator prompting} are expected to be achieved in order, the sub-goals are trained in phases. As detailed in Algorithm \ref{alg:llm_subgoal_reward}, a PPO agent is launched for each reward candidate with sub-goal bonuses increasing ten-fold per stage. Crucially, agents receive no reward for progress beyond the current sub-goal in focus. The \textit{CheckEffectSatisfied} function verifies that an effect is satisfied without unintended changes to other entities. Once a phase is complete, rewards for the subsequent sub-goal are unlocked. When an effect is not achieved, the agent receives a progress reward from the LLM-generated function, scaled by a factor $\textit{max\_reward\_shaping}$ that increases ten-fold per sub-goal. This results in exponentially larger rewards as the agent advances. For instance, in the \textit{grasped-drawer} phase, the agent receives an LLM-shaping reward in $[-1, 1]$ or a $10$ point sub-goal bonus. In the subsequent \textit{open-drawer} phase, the shaping reward scales to $[-10, 10]$, with the final sub-goal bonus reaching $100$.

Upon completing a training phase, the best-performing candidate and its policy are retained. For the subsequent sub-goal, three new candidates are generated; these reuse the ``best'' logic from previous phases (e.g., the optimal grasp-drawer-handle snippet) while providing diverse new logic for the current sub-goal (e.g., open-drawer). We currently use a rule-based function with ground-truth state to verify effects to decouple planning performance from perception errors. However, real-robot deployment would interface with a perception module for grounding. Episode horizons start at 100 steps, increasing by 50 per subsequent stage.
\section{Evaluation}

\begin{table}[th]
\centering
\addtolength{\tabcolsep}{-2pt} 
\vspace*{0.1cm}
\caption{Diversity in Generated Reward Function Candidates ($Temperature=1$). Values denote the number of semantically unique candidates generated per sub-goal.}
\label{tab:reward_diversity}
\footnotesize 
\begin{tabular}{@{}ll@{}}
\toprule
\textbf{Operator} & \textbf{Sub-goals (Number of Variants)} \\ \midrule
pick-up-lid-from-pot & \makecell[l]{grasped\_lid1\_gripper1 (1) \\ not\_(ontop\_lid1\_pot1) (2) \\ not\_(covered\_pot1) (2)} \\ \addlinespace
pick-up-nut-from-peg & \makecell[l]{grasped\_square-nut1\_gripper1 (1) \\ not\_(on-peg\_square-nut1\_round-peg1) (2)} \\ \addlinespace
pick-up-from-open-box & \makecell[l]{grasped\_coffee-pod1\_gripper1 (1) \\ not\_(in\_coffee-pod1\_box1) (1)} \\ \addlinespace
open-drawer & \makecell[l]{grasped\_drawer1\_gripper1 (1) \\ open\_drawer1 (2)} \\ \addlinespace
pick-up-from-open-drawer & \makecell[l]{grasped\_coffee-pod1\_gripper1 (1) \\ not\_(in\_coffee-pod1\_drawer1) (2)} \\ \bottomrule
\end{tabular}
\end{table}

\begin{table}[th]
\centering
\caption{Wilcoxon signed tests p-values for Success Rate and Progress. Comparisons are made between our LLM-Guided (LG) Subgoal agent and the baselines-the LEAGUE-sparse agent (LS) and the Reward Machine (RM). Values marked with * indicate statistical significance ($p < 0.05$).}
\label{tab:p_values}
\begin{tabular}{@{}l cc@{}}
    \toprule
    \multicolumn{3}{c}{\textbf{Success Rate p-values}} \\
    \midrule
    Operator & LS vs LG (Ours) & RM vs LG (Ours)\\
    \midrule
    pick-up-lid-from-pot      & 0.0010* & 0.1094  \\
    pick-up-nut-from-peg      & 0.0010* & 0.0107* \\
    pick-up-from-open-box     & 0.0010* & 0.0010* \\
    open-drawer               & 0.0020* & 0.0010* \\
    pick-up-from-open-drawer  & 0.0010* & 0.0010* \\
    \midrule
    \multicolumn{3}{c}{\textbf{Progress p-values}} \\
    \midrule
    Operator & LS vs LG (Ours) & RM vs LG (Ours)\\
    \midrule
    pick-up-lid-from-pot      & 0.0010* & 0.1094 \\
    pick-up-nut-from-peg      & 0.0010* & 0.0107* \\
    pick-up-from-open-box     & 0.0010* & 0.0010* \\
    open-drawer               & 0.0020* & 0.0010* \\
    pick-up-from-open-drawer  & 0.0010* & 0.0010* \\
    \bottomrule
\end{tabular}

\end{table}
We compare our hybrid planner with the Operator Discovery (OD) method. After encountering a planning impasse, many of the existing hybrid planning and RL approaches launch RL agents to discover a solution. As the RL agent interacts with the environment, state changes are tracked. If the RL agent discovers a plannable state, the state change trace can be used to identify new operators. As shown in Figure \ref{fig:hybrid planner}, we evaluate our approach across four domains of increasing difficulty:
\begin{itemize}
    \item \textbf{Kitchen (Easy):} A novel lid blocks the pot. This domain is considered easy because an RL agent can displace the lid through random exploration to discover a plannable state.
    \item \textbf{Nut Assembly (Medium):} A square nut is placed on a novel round peg. Since the robot's existing operators only support tabletop retrieval, it must learn a new pick-up skill for the peg.
    \item \textbf{Coffee-Box (Medium):} The pod spawns inside a novel box. The agent must learn to retrieve the pod from this container rather than the tabletop.
    \item \textbf{Coffee-Drawer (Hard):} The pod is inside a closed drawer. This is the most difficult task, as random exploration is virtually certain to fail at opening the drawer and retrieving the pod.
\end{itemize}
We launch 10 seeds of Operator Discovery per domain as PPO RL agents. Each agent is capped at 1 million timesteps. We terminate the episode once the agent finds a plannable state. As a comparison, we run our hybrid planner 10 times per domain with GPT-o3 with a temperature of 0.3 to identify the missing operators. We compare our LLM guided sub-goal learning agent with a Reward Machine (RM) baseline that only receives sub-goal bonuses and a LEAGUE-sparse baseline that only receives a reward once all effects are satisfied. We cap the episode length at 100 for the operator learning agent. All experiments are run on a ubuntu 20.04 machine with 24 12th Gen Intel(R) Core(TM) i9-12900K CPUs and a NVIDIA GeForce RTX 4090 GPU. We use the default hyperparameter values for all of our PPO agents.
\section{Discussion}
Table \ref{tab:planner vs rl} shows that the hybrid planner is able to identify the missing operator(s) in all domains. The OD agent can only consistently discover a plannable state in the easy kitchen domain. This shows that OD agents' discovery of missing operators in continuous domains is purely accidental. Therefore OD agents are not a reliable solution to the missing operator identification problem. Figure \ref{fig:sr pr barplots} shows that the LLM guided sub-goal learning agent reaches above 90\% average success and progress for most operators. Wilcoxon signed-tests show that the LLM guided agents significantly outperform the baselines for most operators ($p<0.05$) in Table \ref{tab:p_values}. Additionally, having multiple reward function candidates mitigates the risks of code errors or poorly performing reward functions over single candidate approaches such as LEAGUE++, as shown in Figure \ref{fig:eval_curves}.

The results suggest that the LLM generated reward shaping functions are beneficial in guiding the agents towards achieving the sub-goals. The RM agent outperforms the LS agent in the easy domain, showing that sub-goal bonuses guide the agent towards the overall goal if the sub-goals are easily discoverable. While we demonstrate our approach in simulation, the architecture is applicable to real-world robotics. An implementation would involve object grounding through an open-vocabulary perception module such as OWLv2 \cite{minderer2023scaling} and real-to-sim to allow the LLM-guided RL agents to learn policies in simulation. Sim-to-real techniques help deploy the learned policies back onto the physical robot.
\section{Conclusion and Future Work}
In conclusion, we show that a hybrid LLM‑symbolic planning framework complemented by LLM‑guided RL provides a path for novelty adaptation in robotics.  We outsource the missing operator identification problem and dense reward shaping to LLMs, leveraging their common sense reasoning and code generation capabilities to circumvent the brittleness of hand‑coded planners and the inefficiency of unguided RL. Our evaluation across increasingly difficult domains indicate that LLM‑informed symbolic planning and learning can generalize to varied novel objects, and that the generated reward functions accelerate learning and improve final performance.  Future work will explore scaling this method to environments with multiple concurrent novelties and missing operators. Additionally, we plan to validate the framework on physical robotic platforms using open-vocabulary perception to evaluate real-world grounding and manipulation. Finally, leveraging vision-language models for autonomous predicate invention remains a promising direction to relax the assumption of predicate completeness and enable the dynamic expansion of both predicates and operators.
\bibliographystyle{IEEEtran}
\bibliography{root}

@article{icarte2022reward,
  title={Reward machines: Exploiting reward function structure in reinforcement learning},
  author={Icarte, Rodrigo Toro and Klassen, Toryn Q and Valenzano, Richard and McIlraith, Sheila A},
  journal={Journal of Artificial Intelligence Research},
  volume={73},
  pages={173--208},
  year={2022}
}

@inproceedings{li2024league++,
  title={League++: Empowering continual robot learning through guided skill acquisition with large language models},
  author={Li, Zhaoyi and Yu, Kelin and Cheng, Shuo and Xu, Danfei},
  booktitle={ICLR 2024 Workshop on Large Language Model (LLM) Agents}
}

@article{Gizzi_2022,
   title={Creative Problem Solving in Artificially Intelligent Agents: A Survey and Framework},
   volume={75},
   ISSN={1076-9757},
   url={http://dx.doi.org/10.1613/jair.1.13864},
   DOI={10.1613/jair.1.13864},
   journal={Journal of Artificial Intelligence Research},
   publisher={AI Access Foundation},
   author={Gizzi, Evana and Nair, Lakshmi and Chernova, Sonia and Sinapov, Jivko},
   year={2022},
   month=nov }

@article{minderer2023scaling,
  title={Scaling open-vocabulary object detection},
  author={Minderer, Matthias and Gritsenko, Alexey and Houlsby, Neil},
  journal={Advances in Neural Information Processing Systems},
  volume={36},
  pages={72983--73007},
  year={2023}
}

@inproceedings{disprod23,
author = {Chatterjee, Palash and Chapagain, Ashutosh and Chen, Weizhe and Khardon, Roni},
title = {DiSProD: differentiable symbolic propagation of distributions for planning},
year = {2023},
isbn = {978-1-956792-03-4},
url = {https://doi.org/10.24963/ijcai.2023/591},
doi = {10.24963/ijcai.2023/591},
booktitle = {Proceedings of the Thirty-Second International Joint Conference on Artificial Intelligence},
articleno = {591},
numpages = {9},
location = {Macao, P.R.China},
series = {IJCAI '23}
}

@misc{schulman2017proximalpolicyoptimizationalgorithms,
      title={Proximal Policy Optimization Algorithms}, 
      author={John Schulman and Filip Wolski and Prafulla Dhariwal and Alec Radford and Oleg Klimov},
      year={2017},
      eprint={1707.06347},
      archivePrefix={arXiv},
      primaryClass={cs.LG},
      url={https://arxiv.org/abs/1707.06347}, 
}

@misc{kojima_large_2023,
	title = {Large Language Models are Zero-Shot Reasoners},
	url = {http://arxiv.org/abs/2205.11916},
	doi = {10.48550/arXiv.2205.11916},
	number = {{arXiv}:2205.11916},
	publisher = {{arXiv}},
	author = {Kojima, Takeshi and Gu, Shixiang Shane and Reid, Machel and Matsuo, Yutaka and Iwasawa, Yusuke},
	urldate = {2025-07-30},
	date = {2023-01-29},
        year = {2023},
	eprinttype = {arxiv},
	eprint = {2205.11916 [cs]},
}

@article{wang_self-consistency_2023,
	title = {{SELF}-{CONSISTENCY} {IMPROVES} {CHAIN} {OF} {THOUGHT} {REASONING} {IN} {LANGUAGE} {MODELS}},
	author = {Wang, Xuezhi and Wei, Jason and Schuurmans, Dale and Le, Quoc and Chi, Ed H and Narang, Sharan and Chowdhery, Aakanksha and Zhou, Denny},
	date = {2023},
        year = {2023},
	langid = {english},
}

@misc{kwon_reward_2023,
	title = {Reward Design with Language Models},
	url = {http://arxiv.org/abs/2303.00001},
	doi = {10.48550/arXiv.2303.00001},
	number = {{arXiv}:2303.00001},
	publisher = {{arXiv}},
	author = {Kwon, Minae and Xie, Sang Michael and Bullard, Kalesha and Sadigh, Dorsa},
	urldate = {2025-07-30},
	date = {2023-02-27},
        year = {2023},
	eprinttype = {arxiv},
	eprint = {2303.00001 [cs]},
}

@misc{xie_text2reward_2024,
	title = {Text2Reward: Reward Shaping with Language Models for Reinforcement Learning},
	url = {http://arxiv.org/abs/2309.11489},
	doi = {10.48550/arXiv.2309.11489},
	shorttitle = {Text2Reward},
	number = {{arXiv}:2309.11489},
	publisher = {{arXiv}},
	author = {Xie, Tianbao and Zhao, Siheng and Wu, Chen Henry and Liu, Yitao and Luo, Qian and Zhong, Victor and Yang, Yanchao and Yu, Tao},
	urldate = {2025-07-30},
	date = {2024-05-25},
        year = {2024},
	eprinttype = {arxiv},
	eprint = {2309.11489 [cs]},
}

@misc{ma_eureka_2024,
	title = {Eureka: Human-Level Reward Design via Coding Large Language Models},
	url = {http://arxiv.org/abs/2310.12931},
	doi = {10.48550/arXiv.2310.12931},
	shorttitle = {Eureka},
	number = {{arXiv}:2310.12931},
	publisher = {{arXiv}},
	author = {Ma, Yecheng Jason and Liang, William and Wang, Guanzhi and Huang, De-An and Bastani, Osbert and Jayaraman, Dinesh and Zhu, Yuke and Fan, Linxi and Anandkumar, Anima},
	urldate = {2025-07-30},
	date = {2024-04-30},
        year = {2024},
	eprinttype = {arxiv},
	eprint = {2310.12931 [cs]},
}

@misc{song_self-refined_2023,
	title = {Self-Refined Large Language Model as Automated Reward Function Designer for Deep Reinforcement Learning in Robotics},
	url = {http://arxiv.org/abs/2309.06687},
	doi = {10.48550/arXiv.2309.06687},
	number = {{arXiv}:2309.06687},
	publisher = {{arXiv}},
	author = {Song, Jiayang and Zhou, Zhehua and Liu, Jiawei and Fang, Chunrong and Shu, Zhan and Ma, Lei},
	urldate = {2025-07-30},
	date = {2023-10-02},
        year = {2023},
	eprinttype = {arxiv},
	eprint = {2309.06687 [cs]},
}

@inproceedings{du_guiding_2023,
	title = {Guiding Pretraining in Reinforcement Learning with Large Language Models},
	url = {https://proceedings.mlr.press/v202/du23f.html},
	eventtitle = {International Conference on Machine Learning},
	pages = {8657--8677},
	booktitle = {Proceedings of the 40th International Conference on Machine Learning},
	publisher = {{PMLR}},
	author = {Du, Yuqing and Watkins, Olivia and Wang, Zihan and Colas, Cédric and Darrell, Trevor and Abbeel, Pieter and Gupta, Abhishek and Andreas, Jacob},
	urldate = {2025-07-30},
	date = {2023-07-03},
        year = {2023},
	langid = {english},
	note = {{ISSN}: 2640-3498},
}

@misc{quartey_exploiting_2024,
	title = {Exploiting Contextual Structure to Generate Useful Auxiliary Tasks},
	url = {http://arxiv.org/abs/2303.05038},
	doi = {10.48550/arXiv.2303.05038},
	number = {{arXiv}:2303.05038},
	publisher = {{arXiv}},
	author = {Quartey, Benedict and Shah, Ankit and Konidaris, George},
	urldate = {2025-07-30},
	date = {2024-04-04},
        year ={2024},
	eprinttype = {arxiv},
	eprint = {2303.05038 [cs]},
}

@article{aeronautiques1998pddl,
  title={Pddl—the planning domain definition language},
  author={Aeronautiques, Constructions and Howe, Adele and Knoblock, Craig and McDermott, ISI Drew and Ram, Ashwin and Veloso, Manuela and Weld, Daniel and Sri, David Wilkins and Barrett, Anthony and Christianson, Dave and others},
  journal={Technical Report, Tech. Rep.},
  year={1998}
}

@book{ghallab_automated_2016,
	edition = {1},
	title = {Automated Planning and Acting},
	rights = {https://www.cambridge.org/core/terms},
	isbn = {978-1-107-03727-4 978-1-139-58392-3},
	url = {https://www.cambridge.org/core/product/identifier/9781139583923/type/book},
	publisher = {Cambridge University Press},
	author = {Ghallab, Malik and Nau, Dana and Traverso, Paolo},
	urldate = {2024-07-22},
	date = {2016-07-31},
        year = {2016},
	langid = {english},
	doi = {10.1017/CBO9781139583923},
}

@article{silver_generalized_2024,
	title = {Generalized Planning in {PDDL} Domains with Pretrained Large Language Models},
	volume = {38},
	rights = {Copyright (c) 2024 Association for the Advancement of Artificial Intelligence},
	issn = {2374-3468},
	url = {https://ojs.aaai.org/index.php/AAAI/article/view/30006},
	doi = {10.1609/aaai.v38i18.30006},
	pages = {20256--20264},
	number = {18},
	journaltitle = {Proceedings of the {AAAI} Conference on Artificial Intelligence},
	author = {Silver, Tom and Dan, Soham and Srinivas, Kavitha and Tenenbaum, Joshua B. and Kaelbling, Leslie and Katz, Michael},
	urldate = {2025-07-28},
	date = {2024-03-24},
        year = {2024},
	langid = {english},
	note = {Number: 18},
}

@misc{liu_llmp_2023,
	title = {{LLM}+P: Empowering Large Language Models with Optimal Planning Proficiency},
	url = {http://arxiv.org/abs/2304.11477},
	shorttitle = {{LLM}+P}}

@misc{dagan_dynamic_2023,
	title = {Dynamic Planning with a {LLM}},
	url = {http://arxiv.org/abs/2308.06391},
	number = {{arXiv}:2308.06391},
	publisher = {{arXiv}},
	author = {Dagan, Gautier and Keller, Frank and Lascarides, Alex},
	urldate = {2023-12-11},
	date = {2023-08-11},
        year = {2023},
	eprinttype = {arxiv},
	eprint = {2308.06391 [cs]},
}

@misc{singh_progprompt_2022,
	title = {{ProgPrompt}: Generating Situated Robot Task Plans using Large Language Models},
	url = {http://arxiv.org/abs/2209.11302},
	doi = {10.48550/arXiv.2209.11302},
	shorttitle = {{ProgPrompt}},
	number = {{arXiv}:2209.11302},
	publisher = {{arXiv}},
	author = {Singh, Ishika and Blukis, Valts and Mousavian, Arsalan and Goyal, Ankit and Xu, Danfei and Tremblay, Jonathan and Fox, Dieter and Thomason, Jesse and Garg, Animesh},
	urldate = {2025-07-29},
	date = {2022-09-22},
        year = {2022},
	eprinttype = {arxiv},
	eprint = {2209.11302 [cs]},
}

@misc{yao_react_2023,
	title = {{ReAct}: Synergizing Reasoning and Acting in Language Models},
	url = {http://arxiv.org/abs/2210.03629},
	doi = {10.48550/arXiv.2210.03629},
	shorttitle = {{ReAct}},
	number = {{arXiv}:2210.03629},
	publisher = {{arXiv}},
	author = {Yao, Shunyu and Zhao, Jeffrey and Yu, Dian and Du, Nan and Shafran, Izhak and Narasimhan, Karthik and Cao, Yuan},
	urldate = {2025-07-29},
	date = {2023-03-10},
        year = {2023},
	eprinttype = {arxiv},
	eprint = {2210.03629 [cs]},
}

@misc{huang_language_2022,
	title = {Language Models as Zero-Shot Planners: Extracting Actionable Knowledge for Embodied Agents},
	url = {http://arxiv.org/abs/2201.07207},
	doi = {10.48550/arXiv.2201.07207},
	shorttitle = {Language Models as Zero-Shot Planners},
	number = {{arXiv}:2201.07207},
	publisher = {{arXiv}},
	author = {Huang, Wenlong and Abbeel, Pieter and Pathak, Deepak and Mordatch, Igor},
	urldate = {2025-07-29},
	date = {2022-03-08},
        year = {2022},
	eprinttype = {arxiv},
	eprint = {2201.07207 [cs]},
}

@misc{wang_describe_2024,
	title = {Describe, Explain, Plan and Select: Interactive Planning with Large Language Models Enables Open-World Multi-Task Agents},
	url = {http://arxiv.org/abs/2302.01560},
	doi = {10.48550/arXiv.2302.01560},
	shorttitle = {Describe, Explain, Plan and Select},
	number = {{arXiv}:2302.01560},
	publisher = {{arXiv}},
	author = {Wang, Zihao and Cai, Shaofei and Chen, Guanzhou and Liu, Anji and Ma, Xiaojian and Liang, Yitao},
	urldate = {2025-07-29},
	date = {2024-07-08},
        year = {2024},
	eprinttype = {arxiv},
	eprint = {2302.01560 [cs]},
}

@misc{huang_inner_2022,
	title = {Inner Monologue: Embodied Reasoning through Planning with Language Models},
	url = {http://arxiv.org/abs/2207.05608},
	doi = {10.48550/arXiv.2207.05608},
	shorttitle = {Inner Monologue},
	number = {{arXiv}:2207.05608},
	publisher = {{arXiv}},
	author = {Huang, Wenlong and Xia, Fei and Xiao, Ted and Chan, Harris and Liang, Jacky and Florence, Pete and Zeng, Andy and Tompson, Jonathan and Mordatch, Igor and Chebotar, Yevgen and Sermanet, Pierre and Brown, Noah and Jackson, Tomas and Luu, Linda and Levine, Sergey and Hausman, Karol and Ichter, Brian},
	urldate = {2025-07-29},
	date = {2022-07-12},
        year = {2022},
	eprinttype = {arxiv},
	eprint = {2207.05608 [cs]},
}

@misc{huang_grounded_2023,
	title = {Grounded Decoding: Guiding Text Generation with Grounded Models for Embodied Agents},
	url = {http://arxiv.org/abs/2303.00855},
	doi = {10.48550/arXiv.2303.00855},
	shorttitle = {Grounded Decoding},
	number = {{arXiv}:2303.00855},
	publisher = {{arXiv}},
	author = {Huang, Wenlong and Xia, Fei and Shah, Dhruv and Driess, Danny and Zeng, Andy and Lu, Yao and Florence, Pete and Mordatch, Igor and Levine, Sergey and Hausman, Karol and Ichter, Brian},
	urldate = {2025-07-29},
	date = {2023-12-11},
        year = {2023},
	eprinttype = {arxiv},
	eprint = {2303.00855 [cs]},
}

@misc{ahn_as_2022,
	title = {Do As I Can, Not As I Say: Grounding Language in Robotic Affordances},
	url = {http://arxiv.org/abs/2204.01691},
	doi = {10.48550/arXiv.2204.01691},
	shorttitle = {Do As I Can, Not As I Say},
	number = {{arXiv}:2204.01691},
	publisher = {{arXiv}},
	author = {Ahn, Michael and Brohan, Anthony and Brown, Noah and Chebotar, Yevgen and Cortes, Omar and David, Byron and Finn, Chelsea and Fu, Chuyuan and Gopalakrishnan, Keerthana and Hausman, Karol and Herzog, Alex and Ho, Daniel and Hsu, Jasmine and Ibarz, Julian and Ichter, Brian and Irpan, Alex and Jang, Eric and Ruano, Rosario Jauregui and Jeffrey, Kyle and Jesmonth, Sally and Joshi, Nikhil J. and Julian, Ryan and Kalashnikov, Dmitry and Kuang, Yuheng and Lee, Kuang-Huei and Levine, Sergey and Lu, Yao and Luu, Linda and Parada, Carolina and Pastor, Peter and Quiambao, Jornell and Rao, Kanishka and Rettinghouse, Jarek and Reyes, Diego and Sermanet, Pierre and Sievers, Nicolas and Tan, Clayton and Toshev, Alexander and Vanhoucke, Vincent and Xia, Fei and Xiao, Ted and Xu, Peng and Xu, Sichun and Yan, Mengyuan and Zeng, Andy},
	urldate = {2025-07-29},
	date = {2022-08-16},
        year = {2022},
	eprinttype = {arxiv},
	eprint = {2204.01691 [cs]},
}

@misc{zhao2023largelanguagemodelscommonsense,
      title={Large Language Models as Commonsense Knowledge for Large-Scale Task Planning}, 
      author={Zirui Zhao and Wee Sun Lee and David Hsu},
      year={2023},
      eprint={2305.14078},
      archivePrefix={arXiv},
      primaryClass={cs.RO},
      url={https://arxiv.org/abs/2305.14078}, 
}

@inproceedings{shukla_lgts_2024,
	location = {Richland, {SC}},
	title = {{LgTS}: Dynamic Task Sampling using {LLM}-generated Sub-Goals for Reinforcement Learning Agents},
	isbn = {9798400704864},
	series = {{AAMAS} '24},
	shorttitle = {{LgTS}},
	pages = {1736--1744},
	booktitle = {Proceedings of the 23rd International Conference on Autonomous Agents and Multiagent Systems},
	publisher = {International Foundation for Autonomous Agents and Multiagent Systems},
	author = {Shukla, Yash and Gao, Wenchang and Sarathy, Vasanth and Velasquez, Alvaro and Wright, Robert and Sinapov, Jivko},
	urldate = {2025-07-27},
	date = {2024-05-06},
        year = {2024}
}

@misc{mandlekar_mimicgen_2023,
	title = {{MimicGen}: A Data Generation System for Scalable Robot Learning using Human Demonstrations},
	url = {http://arxiv.org/abs/2310.17596},
	doi = {10.48550/arXiv.2310.17596},
	shorttitle = {{MimicGen}},
	number = {{arXiv}:2310.17596},
	publisher = {{arXiv}},
	author = {Mandlekar, Ajay and Nasiriany, Soroush and Wen, Bowen and Akinola, Iretiayo and Narang, Yashraj and Fan, Linxi and Zhu, Yuke and Fox, Dieter},
	urldate = {2025-07-28},
	date = {2023-10-26},
        year = {2023},
	langid = {english},
	eprinttype = {arxiv},
	eprint = {2310.17596 [cs]},
}

@misc{sarathy_spotter_2020,
	title = {{SPOTTER}: Extending Symbolic Planning Operators through Targeted Reinforcement Learning},
	url = {http://arxiv.org/abs/2012.13037},
	doi = {10.48550/arXiv.2012.13037},
	shorttitle = {{SPOTTER}},
	number = {{arXiv}:2012.13037},
	publisher = {{arXiv}},
	author = {Sarathy, Vasanth and Kasenberg, Daniel and Goel, Shivam and Sinapov, Jivko and Scheutz, Matthias},
	urldate = {2025-07-28},
	date = {2020-12-24},
        year = {202},
	eprinttype = {arxiv},
	eprint = {2012.13037 [cs]},
}

@misc{lorang_curiosity-driven_2025,
	title = {Curiosity-Driven Imagination: Discovering Plan Operators and Learning Associated Policies for Open-World Adaptation},
	url = {http://arxiv.org/abs/2503.04931},
	doi = {10.48550/arXiv.2503.04931},
	shorttitle = {Curiosity-Driven Imagination},
	number = {{arXiv}:2503.04931},
	publisher = {{arXiv}},
	author = {Lorang, Pierrick and Lu, Hong and Scheutz, Matthias},
	urldate = {2025-07-28},
	date = {2025-03-06},
        year = {2025},
	eprinttype = {arxiv},
	eprint = {2503.04931 [cs]},
}

@article{goel_neurosymbolic_2024,
	title = {A neurosymbolic cognitive architecture framework for handling novelties in open worlds},
	volume = {331},
	issn = {0004-3702},
	url = {https://www.sciencedirect.com/science/article/pii/S000437022400047X},
	doi = {10.1016/j.artint.2024.104111},
	pages = {104111},
	journaltitle = {Artificial Intelligence},
	shortjournal = {Artificial Intelligence},
	author = {Goel, Shivam and Lymperopoulos, Panagiotis and Thielstrom, Ravenna and Krause, Evan and Feeney, Patrick and Lorang, Pierrick and Schneider, Sarah and Wei, Yichen and Kildebeck, Eric and Goss, Stephen and Hughes, Michael C. and Liu, Liping and Sinapov, Jivko and Scheutz, Matthias},
	urldate = {2025-07-28},
	date = {2024-06-01},
        year = {2024},
}

@misc{wei_chain--thought_2023,
	title = {Chain-of-Thought Prompting Elicits Reasoning in Large Language Models},
	url = {http://arxiv.org/abs/2201.11903},
	number = {{arXiv}:2201.11903},
	publisher = {{arXiv}},
	author = {Wei, Jason and Wang, Xuezhi and Schuurmans, Dale and Bosma, Maarten and Ichter, Brian and Xia, Fei and Chi, Ed and Le, Quoc and Zhou, Denny},
	urldate = {2023-10-28},
	date = {2023-01-10},
        year = {2023},
	eprinttype = {arxiv},
	eprint = {2201.11903 [cs]},
}

@inproceedings{lorang_framework_2024,
	title = {A Framework for Neurosymbolic Goal-Conditioned Continual Learning in Open World Environments},
	url = {https://ieeexplore.ieee.org/abstract/document/10801627},
	doi = {10.1109/IROS58592.2024.10801627},
	eventtitle = {2024 {IEEE}/{RSJ} International Conference on Intelligent Robots and Systems ({IROS})},
	pages = {12070--12077},
	booktitle = {2024 {IEEE}/{RSJ} International Conference on Intelligent Robots and Systems ({IROS})},
	author = {Lorang, Pierrick and Goel, Shivam and Shukla, Yash and Zips, Patrik and Scheutz, Matthias},
	urldate = {2025-07-28},
	date = {2024-10},
        year = {2024},
	note = {{ISSN}: 2153-0866},
}

@inproceedings{lorang_adapting_2024,
	title = {Adapting to the “Open World”: The Utility of Hybrid Hierarchical Reinforcement Learning and Symbolic Planning},
	url = {https://ieeexplore.ieee.org/abstract/document/10611594},
	doi = {10.1109/ICRA57147.2024.10611594},
	shorttitle = {Adapting to the “Open World”},
	eventtitle = {2024 {IEEE} International Conference on Robotics and Automation ({ICRA})},
	pages = {508--514},
	booktitle = {2024 {IEEE} International Conference on Robotics and Automation ({ICRA})},
	author = {Lorang, Pierrick and Horvath, Helmut and Kietreiber, Tobias and Zips, Patrik and Heitzinger, Clemens and Scheutz, Matthias},
	urldate = {2025-07-28},
	date = {2024-05},
        year = {2024},
}

@inproceedings{goel_rapid-learn_2022,
	title = {{RAPid}-Learn: A Framework for Learning to Recover for Handling Novelties in Open-World Environments},
	url = {https://ieeexplore.ieee.org/abstract/document/9962230},
	doi = {10.1109/ICDL53763.2022.9962230},
	shorttitle = {{RAPid}-Learn},
	eventtitle = {2022 {IEEE} International Conference on Development and Learning ({ICDL})},
	pages = {15--22},
	booktitle = {2022 {IEEE} International Conference on Development and Learning ({ICDL})},
	author = {Goel, Shivam and Shukla, Yash and Sarathy, Vasanth and Scheutz, Matthias and Sinapov, Jivko},
	urldate = {2025-07-28},
	date = {2022-09},
        year = {2022},
}

@misc{yang_peorl_2018,
	title = {{PEORL}: Integrating Symbolic Planning and Hierarchical Reinforcement Learning for Robust Decision-Making},
	url = {http://arxiv.org/abs/1804.07779},
	doi = {10.48550/arXiv.1804.07779},
	shorttitle = {{PEORL}},
	number = {{arXiv}:1804.07779},
	publisher = {{arXiv}},
	author = {Yang, Fangkai and Lyu, Daoming and Liu, Bo and Gustafson, Steven},
	urldate = {2025-07-28},
	date = {2018-06-05},
        year = {2018},
	eprinttype = {arxiv},
	eprint = {1804.07779 [cs]},
}

@article{cheng_league_2023, title={LEAGUE: Guided Skill Learning and Abstraction for Long-Horizon Manipulation}, volume={8}, ISSN={2377-3766}, DOI={10.1109/LRA.2023.3308061}, abstractNote={To assist with everyday human activities, robots must solve complex long-horizon tasks and generalize to new settings. Recent deep reinforcement learning (RL) methods show promise in fully autonomous learning, but they struggle to reach long-term goals in large environments. On the other hand, Task and Motion Planning (TAMP) approaches excel at solving and generalizing across long-horizon tasks, thanks to their powerful state and action abstractions. But they assume predefined skill sets, which limits their real-world applications. In this work, we combine the benefits of these two paradigms and propose an integrated task planning and skill learning framework named LEAGUE (Learning and Abstraction with Guidance). LEAGUE leverages the symbolic interface of a task planner to guide RL-based skill learning and creates abstract state space to enable skill reuse. More importantly, LEAGUE learns manipulation skills in-situ of the task planning system, continuously growing its capability and the set of tasks that it can solve. We evaluate LEAGUE on four challenging simulated task domains and show that LEAGUE outperforms baselines by large margins. We also show that the learned skills can be reused to accelerate learning in new tasks domains and transfer to a physical robot platform.}, number={10}, journal={IEEE Robotics and Automation Letters}, author={Cheng, Shuo and Xu, Danfei}, year={2023}, month=oct, pages={6451–6458} }

@misc{silver_learning_2022,
	title = {Learning Neuro-Symbolic Skills for Bilevel Planning},
	url = {http://arxiv.org/abs/2206.10680},
	doi = {10.48550/arXiv.2206.10680},
	number = {{arXiv}:2206.10680},
	publisher = {{arXiv}},
	author = {Silver, Tom and Athalye, Ashay and Tenenbaum, Joshua B. and Lozano-Perez, Tomas and Kaelbling, Leslie Pack},
	urldate = {2025-07-28},
	date = {2022-10-12},
        year ={2022},
	eprinttype = {arxiv},
	eprint = {2206.10680 [cs]},
}

@article{kokel_reprel_2021,
	title = {{RePReL}: Integrating Relational Planning and Reinforcement Learning for Effective Abstraction},
	volume = {31},
	rights = {Copyright (c) 2021 Association for the Advancement of Artificial Intelligence},
	issn = {2334-0843},
	url = {https://ojs.aaai.org/index.php/ICAPS/article/view/16001},
	doi = {10.1609/icaps.v31i1.16001},
	shorttitle = {{RePReL}},
	pages = {533--541},
	journaltitle = {Proceedings of the International Conference on Automated Planning and Scheduling},
	author = {Kokel, Harsha and Manoharan, Arjun and Natarajan, Sriraam and Ravindran, Balaraman and Tadepalli, Prasad},
	urldate = {2025-07-28},
	date = {2021-05-17},
        year = {2021},
	langid = {english},
}

@article{illanes_symbolic_2020,
	title = {Symbolic Plans as High-Level Instructions for Reinforcement Learning},
	volume = {30},
	rights = {Copyright (c) 2020 Association for the Advancement of Artificial Intelligence},
	issn = {2334-0843},
	url = {https://ojs.aaai.org/index.php/ICAPS/article/view/6750},
	doi = {10.1609/icaps.v30i1.6750},
	pages = {540--550},
	journaltitle = {Proceedings of the International Conference on Automated Planning and Scheduling},
	author = {Illanes, León and Yan, Xi and Icarte, Rodrigo Toro and {McIlraith}, Sheila A.},
	urldate = {2025-07-28},
	date = {2020-06-01},
        year = {2020},
	langid = {english},
}

@article{acharya_neurosymbolic_2024,
	title = {Neurosymbolic Reinforcement Learning and Planning: A Survey},
	volume = {5},
	issn = {2691-4581},
	url = {https://ieeexplore.ieee.org/document/10238788},
	doi = {10.1109/TAI.2023.3311428},
	shorttitle = {Neurosymbolic Reinforcement Learning and Planning},
	pages = {1939--1953},
	number = {5},
	journaltitle = {{IEEE} Transactions on Artificial Intelligence},
	author = {Acharya, Kamal and Raza, Waleed and Dourado, Carlos and Velasquez, Alvaro and Song, Houbing Herbert},
	urldate = {2025-07-28},
	date = {2024-05},
        year = {2024},
}

@misc{balloch_neuro-symbolic_2023,
	title = {Neuro-Symbolic World Models for Adapting to Open World Novelty},
	url = {http://arxiv.org/abs/2301.06294},
	doi = {10.48550/arXiv.2301.06294},
	number = {{arXiv}:2301.06294},
	publisher = {{arXiv}},
	author = {Balloch, Jonathan and Lin, Zhiyu and Wright, Robert and Peng, Xiangyu and Hussain, Mustafa and Srinivas, Aarun and Kim, Julia and Riedl, Mark O.},
	urldate = {2025-07-28},
	date = {2023-01-16},
        year = {2023},
	eprinttype = {arxiv},
	eprint = {2301.06294 [cs]},
}

@inproceedings{guan_leveraging_2022,
	title = {Leveraging Approximate Symbolic Models for Reinforcement Learning via Skill Diversity},
	url = {https://proceedings.mlr.press/v162/guan22c.html},
	eventtitle = {International Conference on Machine Learning},
	pages = {7949--7967},
	booktitle = {Proceedings of the 39th International Conference on Machine Learning},
	publisher = {{PMLR}},
	author = {Guan, Lin and Sreedharan, Sarath and Kambhampati, Subbarao},
	urldate = {2025-07-28},
	date = {2022-06-28},
        year = {2022},
	langid = {english},
	note = {{ISSN}: 2640-3498},
}

@article{stable-baselines3,
  author  = {Antonin Raffin and Ashley Hill and Adam Gleave and Anssi Kanervisto and Maximilian Ernestus and Noah Dormann},
  title   = {Stable-Baselines3: Reliable Reinforcement Learning Implementations},
  journal = {Journal of Machine Learning Research},
  year    = {2021},
  volume  = {22},
  number  = {268},
  pages   = {1-8},
  url     = {http://jmlr.org/papers/v22/20-1364.html}
}

\end{document}